\documentclass[10pt, a4paper]{article}
\usepackage[final]{lrec2026} 
\title{From Consensus to Split Decisions: ABC-Stratified Sentiment in Holocaust Oral Histories}
\name{Daban Q. Jaff} 
\address{
  Erfurt University, Erfurt, Germany \\
  daban.hamad\_ameen@uni-erfurt.de
}
\abstract{
Polarity detection becomes substantially more challenging under domain shift, particularly in heterogeneous, long-form narratives with complex discourse structure, such as Holocaust oral histories. This paper presents a corpus-scale diagnostic study of off-the-shelf sentiment classifiers on long-form Holocaust oral histories, using three pretrained transformer-based polarity classifiers on a corpus of 107{,}305 utterances and 579{,}013 sentences. After assembling model outputs, we introduce an agreement-based stability taxonomy (ABC) to stratify inter-model output stability. We report pairwise percent agreement, Cohen’s $\kappa$, Fleiss’ $\kappa$, and row-normalized confusion matrices to localize systematic disagreement. As an auxiliary descriptive signal, a T5-based emotion classifier is applied to stratified samples from each agreement stratum to compare emotion distributions across strata. The combination of multi-model label triangulation and the ABC taxonomy provides a cautious, operational framework for characterizing where and how sentiment models diverge in sensitive historical narratives. Inter-model agreement is low to moderate overall and is driven primarily by boundary decisions around neutrality.
}
\usepackage{amsmath, amssymb, booktabs, tabularx, array, graphicx, placeins, url}
\usepackage{stfloats}
\usepackage{xurl}
\newcolumntype{L}{>{\raggedright\arraybackslash}X}

\begin{document}
\maketitleabstract

\section{Introduction}
Sentiment analysis (SA) focuses on identifying evaluative meanings, such as polarity or emotional states \cite{cambria2017sentiment}. It is typically framed as a component of opinion mining, where the goal is to extract attitudes toward specific entities or events \cite{pang2008opinion, liu2012sentiment}. Methodologically, SA has evolved from lexicon-based \cite{taboada2011lexicon} and rule-based \cite{hutto2014vader} to machine learning \cite{turney2002thumbs} and modern transformer architectures \cite{vaswani2017attention, devlin2019bert}. The contemporary models leverage deep contextual representations, though their effectiveness remains heavily dependent on the chosen unit of analysis, especially on whether it is a single sentence or an entire document \cite{pang2008opinion, liu2012sentiment}.

A major obstacle for applying off-the-shelf SA systems to Holocaust oral histories is domain shift: models trained on different genres (e.g., product reviews or Twitter) face a changed input distribution when applied to long-form historical narratives, which can alter label propensities \cite{blitzer2007domain, glorot2011domain, pan2010transfer}. In Holocaust oral histories, evaluative meaning is often expressed indirectly (through description of the lived experience, stance taking, or moral framing), distributed across multiple sentences, and confounded by reported speech and the verbal reconstruction of experience over time. These characteristics can reduce the density of explicit sentiment cues and make polarity judgments less stable.

This paper studies the resulting phenomenon of model disagreement in Holocaust oral histories. We run three pretrained sentiment models and quantify how strongly they disagree at both sentence and utterance levels. The central methodological goal is neither to identify a single best model nor to estimate sentiment accuracy against human-annotated ground truth, but rather to harness the heterogeneous knowledge and inductive biases of these three systems simultaneously. None of the models were fine-tuned on trauma-related discourse. By treating each classifier as an independent knowledge source shaped by its training distribution, the pipeline is designed to expose genuine domain-shift behavior. Moreover, model confidences are not calibrated and are not directly comparable across architectures or training regimes; they are used here only as within-model heuristics and descriptive proxies.

\section{Related Work}
\label{sec:related}
SA is known to degrade under domain shift and in long-form narrative settings with domain-specific linguistic phenomena. Early work shows substantial performance drops when sentiment models are transferred across domains \citep{blitzer2007domain}, and transfer-learning surveys attribute this to distribution mismatch between source and target data \citep{pan2010transfer}. Domain variation in vocabulary and expression is therefore a core obstacle for opinion mining \citep{liu2012sentiment}, motivating domain-adaptation approaches that jointly model diverse domains \citep{barnes2018projecting}.

Sentiment inference also depends on task definition and textual structure. Unit choice matters because document-level sentiment is not simply an average of sentence sentiments \citep{pang2008opinion}, and aggregation interacts with how opinions are expressed across discourse \citep{liu2012sentiment,kraus2019discourse}. Moreover, subjectivity and attribution can confound polarity when evaluations are embedded in reported or narrative speech \citep{wilson2005attributions,wiebe2005opinions_emotions}. To address these issues in practice, ensembling polarities across models is commonly used to combine complementary sentiment systems, including confidence averaging, stacking, and neural ensembles \citep{hagen2015twitter,troncy2017sentime,rouvier2017lia}. 

There is emerging computational work on sentiment, emotion, and text classification in oral-history interviews. Recent examples include neural sentiment analysis on Holocaust interviews \citep{blanke2020}, emotion recognition in German oral histories \citep{gref2022}, geographic emotion modeling of Holocaust testimonies \citep{ezeani2024}, emotion annotation in the ACT UP Oral History Project \citep{pessanha2025}, and LLM-based classification of Japanese-American incarceration narratives \citep{chen2024,cherukuri2025}. 

Unlike prior works, this study focuses on systematic inter-model disagreement in Holocaust oral histories. To do so, we introduce an ABC agreement taxonomy (with an A split used for polarity-specific analyses) and analyze agreement/divergence across sentence- and utterance-level predictions.

\section{Method: Triangulation and ABC Taxonomy}
We employ three off-the-shelf pretrained transformer sentiment classifiers: SiEBERT \citep{hartmann2023more}, CardiffNLP Twitter-RoBERTa \citep{barbieri2020tweeteval}, and NLPTown (nlptown/bert-base-multilingual-uncased-sentiment). The models are deliberately selected to capture the complementary knowledge encoded in models trained under markedly different regimes: general web text, Twitter-style conversational language, and multilingual product reviews. Moreover, SiEBERT was included intentionally despite its binary label space because its forced polarity decisions make unanimous agreement more conservative and make disagreement with the Neutral-capable models analytically useful under domain shift.

Each utterance \(u\) is segmented into sentences using NLTK’s \texttt{sent\_tokenize} (punkt-based). Only minimal normalization (whitespace cleanup) is applied prior to segmentation. Label harmonization follows the upstream pipelines: \textsc{NLPTown}’s 1--5 star ratings are mapped to three-way polarity (1--2: \textsc{Negative}, 3: \textsc{Neutral}, 4--5: \textsc{Positive}) (confidence reflects certainty in the winning star rating).

For each sentence, each model outputs a label and an associated confidence score
(the model's predicted probability for that label). For each model, we additionally
compute an utterance-level aggregated label by assigning each polarity label $\ell$
the score
\[
s(\ell) = \frac{n_\ell}{N}\,\operatorname{meanConf}(\ell),
\]
where $n_\ell$ is the number of sentence-level outputs assigned label $\ell$ and
$N$ is the total number of sentences in the utterance. We then select the label
$\operatorname*{arg\,max}_{\ell} s(\ell)$. Scripts and analysis materials are publicly available.\footnote{\url{https://github.com/dabjaff/ABC-Stratified-Sentiment-in-Holocaust-Oral-Histories}.}

Table~\ref{tab:polarity_dist} summarizes the marginal polarity distributions produced by each model at sentence and utterance levels. These distributions reveal substantial label-propensity differences across models. This motivates the agreement diagnostics and ABC stratification introduced below.

\begin{table}[t]
\centering
\small
\setlength{\tabcolsep}{4pt}
\renewcommand{\arraystretch}{1.05}
\begin{tabularx}{\columnwidth}{@{} l @{\hspace{8pt}} L r r r @{}}
\toprule
\textbf{Level} & \textbf{Model} & \textbf{Neg.\ (\%)} & \textbf{Neu.\ (\%)} & \textbf{Pos.\ (\%)} \\
\midrule
\multicolumn{5}{@{}l}{\textit{\shortstack[l]{Utterance\\$107{,}305$}}} \\
& \textsc{NLPTown} & 48.2 & 19.2 & 32.6 \\
& \textsc{CardiffNLP} & 11.3 & 80.4 & 8.3 \\
& \textsc{SiEBERT}  & 54.2 & --- & 45.8 \\
\addlinespace
\multicolumn{5}{@{}l}{\textit{\shortstack[l]{Sentence\\$579{,}013$}}} \\
& \textsc{NLPTown} & 45.1 & 24.0 & 31.0 \\
& \textsc{CardiffNLP} & 21.2 & 69.3 & 9.4 \\
& \textsc{SiEBERT} & 53.9 & --- & 46.1 \\
\bottomrule
\end{tabularx}
\caption{Polarity distributions across models and granularities.}
\label{tab:polarity_dist}
\end{table}

\subsection{Triangulation}
After obtaining sentence-level labels and confidence scores from all three models, and deriving one aggregated utterance-level label per model, we perform cross-model triangulation at two granularities (Table~\ref{tab:Triangulation}), as follows. 

\begin{table}[h]
\centering
\small
\setlength{\tabcolsep}{4pt}
\renewcommand{\arraystretch}{1.1}
\begin{tabularx}{\columnwidth}{@{} l @{\hspace{12pt}} X r r r @{}}
\toprule
\textbf{Level} & \textbf{Metric} & \textbf{Neg. (\%)} & \textbf{Neu. (\%)} & \textbf{Pos. (\%)} \\
\midrule
\multicolumn{5}{@{}l}{\textit{\shortstack[l]{Utterance\\$107{,}305$}}} \\
& Count & 49,133 & 18,162 & 40,009 \\
& Percentage & 45.8\% & 16.9\% & 37.3\% \\
\addlinespace
\multicolumn{5}{@{}l}{\textit{\shortstack[l]{Sentence\\$579{,}013$}}} \\
& Count & 260,727 & 113,237 & 205,049 \\
& Percentage & 45.0\% & 19.6\% & 35.4\% \\
\bottomrule
\end{tabularx}
\caption{Triangulated polarity distribution across granularities.}
\label{tab:Triangulation}
\end{table}

\paragraph{Sentence level.}
A consensus label is obtained by majority vote across the three models. If at least two models agree, that label is selected. In the case of a true three-way split (one \textsc{Negative}, one \textsc{Neutral}, one \textsc{Positive}), the label with the highest model-reported confidence is selected.

\paragraph{Utterance level.}
For each model separately, we first use the utterance-level aggregation procedure defined above to obtain one aggregated polarity label from that model's sentence-level outputs. Cross-model triangulation at the utterance level then applies majority vote over these three model-specific aggregated labels (equivalently represented as $-1,0,+1$ for analysis); sentence-level labels and confidence scores are not consulted directly at this stage. In rare triangulation edge cases requiring deterministic tie resolution (95 sentences; 16 utterances), \textsc{SiEBERT} is used as a fallback to ensure reproducible label assignment, not as a claim of superior validity.

\subsubsection{Stability Stratification: ABC Taxonomy}
While triangulation produces an ensemble label at both granularities, it does not by itself indicate label stability, i.e., the degree of inter-model agreement or disagreement associated with that label. We therefore introduce the ABC taxonomy as a diagnostic stratification framework that tags the outputs of cross-model triangulation by inter-model agreement stability. In this framework, each category represents a different level of consensus across the three-model ensemble, and each sentence and utterance is assigned to one of three agreement categories (see Table~\ref{tab:abc_metrics}).

\begin{itemize}
\item \textbf{Category A (Full Agreement)}: All three models assign the exact same polarity, and the shared polarity is either \textsc{Positive} or \textsc{Negative}, because \textsc{SiEBERT} does not produce a Neutral class.
\item \textbf{Category B (Partial Agreement)}: Exactly two models agree on the label. This includes (i) cases where the agreement involves \textsc{Neutral} (from \textsc{CardiffNLP} or \textsc{NLPTown}) and (ii) cases where at least two models agree on \textsc{Positive} or \textsc{Negative}.
\item \textbf{Category C (Maximal Conflict)}: The three models produce three distinct labels (one \textsc{Negative}, one \textsc{Neutral}, one \textsc{Positive}).
\end{itemize}
Because \textsc{SiEBERT} cannot emit \textsc{Neutral}, Category~A should be interpreted as a conservative unanimity subset for non-neutral polarity only, not as a general high-reliability subset over the full three-way label space.
\begin{table}[h]
\centering
\small
\setlength{\tabcolsep}{6pt}
\renewcommand{\arraystretch}{1.10}
\begin{tabular*}{\columnwidth}{@{\extracolsep{\fill}} l l r r @{}}
\toprule
\textbf{Level} & \textbf{Cat.} & \textbf{Count ($n$)} & \textbf{Share (\%)} \\
\midrule
\multicolumn{4}{@{}l}{\textit{U ($N=107{,}305$)}} \\
& A$_{-1}$ & 8{,}786 & 8.2 \\
& A$_{+1}$ & 6{,}873 & 6.4 \\
& B & 73{,}037 & 68.1 \\
& C & 18{,}609 & 17.3 \\
\addlinespace
\multicolumn{4}{@{}l}{\textit{S ($N=579{,}013$)}} \\
& A$_{-1}$ & 85{,}372 & 14.7 \\
& A$_{+1}$ & 39{,}119 & 6.8 \\
& B & 365{,}243 & 63.1 \\
& C & 89{,}279 & 15.4 \\
\bottomrule
\end{tabular*}
\caption{ABC taxonomy prevalence by granularity.}
\label{tab:abc_metrics}
\end{table}

\subsection{Kappa-based Agreement Diagnostics}
To quantify agreement under model variation and complement the discrete agreement strata (A/B/C), we report standard inter-rater reliability diagnostics by treating the three sentiment models as raters and each sentence/utterance as an item (Table~\ref{tab:kappa_metrics}).

\begin{table}[h]
\centering
\scriptsize
\setlength{\tabcolsep}{2.8pt}
\renewcommand{\arraystretch}{1.05}
\begin{tabularx}{\columnwidth}{@{} L r r r r r @{}}
\toprule
\textbf{Pair} & \textbf{Agr} & \textbf{$\kappa$} &
\textbf{$N_{\neq 0}$} & \textbf{Agr$_{\neq 0}$} & \textbf{$\kappa_{\neq 0}$} \\
\midrule
\multicolumn{6}{@{}l}{\textit{Utterances} ($N{=}107{,}305$)} \\
\textsc{SiEBERT}--\textsc{CardiffNLP}   & 17.8 & 0.088 & 21{,}045  & 90.9 & 0.816 \\
\textsc{SiEBERT}--\textsc{NLPTown}      & 61.7 & 0.350 & 86{,}746  & 76.3 & 0.518 \\
\textsc{CardiffNLP}--\textsc{NLPTown}   & 32.3 & 0.114 & 18{,}649  & 88.5 & 0.767 \\
\addlinespace
\multicolumn{6}{@{}l}{\textit{Sentences} ($N{=}579{,}013$)} \\
\textsc{SiEBERT}--\textsc{CardiffNLP}   & 28.0 & 0.144 & 177{,}588 & 91.2 & 0.801 \\
\textsc{SiEBERT}--\textsc{NLPTown}      & 57.5 & 0.308 & 440{,}159 & 75.6 & 0.504 \\
\textsc{CardiffNLP}--\textsc{NLPTown}   & 42.1 & 0.184 & 150{,}755 & 87.5 & 0.719 \\
\bottomrule
\end{tabularx}
\caption{Pairwise agreement (Agr.) and $\kappa$ for 3-way and polarity-only ($N_{\neq 0}$) subsets.}
\label{tab:kappa_metrics}
\end{table}
For each model pair, we compute percent agreement (\textbf{Agr.}) and Cohen's $\kappa$ on the shared three-way label space (\textsc{Negative}/\textsc{Neutral}/\textsc{Positive}). Because \textsc{SiEBERT} is binary while \textsc{CardiffNLP} and \textsc{NLPTown} are three-class, Neutral-inclusive agreement and confusion patterns are not directly comparable across all model pairs. We therefore interpret Neutral-boundary effects primarily in the pair where both models can emit \textsc{Neutral} (\textsc{CardiffNLP} and \textsc{NLPTown}). In addition, we exclude any unit labeled \textsc{Neutral} by either model in a given pair and recompute agreement and $\kappa_{\neq 0}$ over \{\textsc{Negative}, \textsc{Positive}\}.

Finally, we compute Fleiss' $\kappa$ (three raters) on the three-way space and on the polarity-only subset to summarize overall agreement, and we produce row-normalized confusion matrices for each classifier pair to localize which labels drive disagreement.

\begin{table}[h]
\centering
\small
\setlength{\tabcolsep}{4pt}
\renewcommand{\arraystretch}{1.05}
\begin{tabularx}{\columnwidth}{@{} l r r r r @{}}
\toprule
\textbf{Level} & \textbf{$N$ (3-way)} & \textbf{Fleiss' $\kappa$} & \textbf{$N_{\neq 0}$} & \textbf{Fleiss' $\kappa_{\neq 0}$} \\
\midrule
Utterance & 107{,}305 & 0.0535 & 18{,}649  & 0.7835 \\
Sentence  & 579{,}013 & 0.1287 & 150{,}755 & 0.7398 \\
\bottomrule
\end{tabularx}
\caption{Overall three-model agreement (Fleiss' $\kappa$) on the full three-way label space and on the polarity-only subset ($N_{\neq 0}$), where units labeled \textsc{Neutral} by \textsc{CardiffNLP} or \textsc{NLPTown} are excluded.}
\label{tab:fleiss_kappa}
\end{table}

\subsection{Auxiliary Emotion Profiling}

As an auxiliary descriptive signal, we apply a T5-based emotion classifier (\texttt{mrm8488/t5-base-finetuned-emotion}; \citealp{raffel2020t5}) to assess whether ABC strata exhibit coherent affective profiles. We use T5 as a descriptive probe because it predicts discrete emotions within a different model family/objective (text-to-text generation), reducing the risk of simply reproducing the same polarity decision boundary. Like the sentiment models, it remains out-of-domain for Holocaust testimony discourse, so its outputs are interpreted descriptively only.

Because Category~A splits into two polarity-consistent subsets, we define four groups for affective triangulation: $A_{+1}$ (full tri-model agreement on \textsc{Positive}), $A_{-1}$ (full tri-model agreement on \textsc{Negative}), and Categories~B and~C. We randomly sample $2{,}000$ utterances ($500$ per group) and $4{,}000$ sentences ($1{,}000$ per group), restricting inputs to 10--350 words. For each group, we compute (i) emotion-label distributions at sentence and utterance levels and (ii) mean confidence for the predicted emotion label (reported as a relative proxy, not a calibrated probability). We then compare these profiles across groups to assess whether agreement strata align with distinct affective signatures.

\subsection{Data}
The pipeline is applied to CORHOH \citep{Jaff2025CORHOH} (see Table~\ref{tab:corpus_demographics}), and only survivor utterances are analyzed (107{,}305 utterances, segmented into 579{,}013 sentences).

\begin{table}[h]
\centering
\small
\setlength{\tabcolsep}{4pt}
\renewcommand{\arraystretch}{1.05}
\begin{tabularx}{\columnwidth}{@{} l L r r @{}}
\toprule
\textbf{Category} & \textbf{Attribute} & \textbf{Count} & \textbf{\%} \\
\midrule
\textbf{Gender}
  & Female & 270 & 54.0 \\
  & Male & 230 & 46.0 \\
\addlinespace
\textbf{Birth cohort}
  & 1890s--1910s & 120 & 24.0 \\
  & 1920s & 320 & 64.0 \\
  & 1930s & 60 & 12.0 \\
\addlinespace
\textbf{Top birthplaces}
  & Poland & 181 & 36.2 \\
  & Germany & 115 & 23.0 \\
  & Other (23 loc.) & 204 & 40.8 \\
\addlinespace
\textbf{Migration era}
  & 1930s--1940s & 273 & 54.6 \\
  & 1950s & 111 & 22.2 \\
  & Other/Unknown & 116 & 23.2 \\
\addlinespace
\bottomrule
\end{tabularx}
\caption{Corpus demographics and background variables (\textit{N=500}).}
\label{tab:corpus_demographics}
\end{table}
\section{Results}
\subsection{Model-wise Polarity Distributions}
Before turning to agreement metrics, Table~\ref{tab:polarity_dist} shows that the three models produce sharply different marginal polarity distributions across both granularities, indicating strong label-propensity mismatch under domain shift. In particular, \textsc{CardiffNLP} is strongly \textsc{Neutral}-dominant, \textsc{NLPTown} is substantially more polar, and \textsc{SiEBERT} is strictly binary; these model-specific output profiles motivate the inter-model diagnostics reported next.

\subsection{Inter-model Classification}

Inter-model classification is examined using pairwise $\kappa$-based diagnostics (Table~\ref{tab:kappa_metrics}), overall three-model agreement via Fleiss' $\kappa$ (Table~\ref{tab:fleiss_kappa}), and row-normalized confusion matrices (Table~\ref{tab:conf_nlptown_card_row}).
\begin{table}[h]
\centering
\scriptsize
\setlength{\tabcolsep}{3pt}
\renewcommand{\arraystretch}{1.05}

\textbf{Sentence-level (N=579{,}013)}\\[-1pt]
\begin{tabularx}{\columnwidth}{l *{3}{>{\centering\arraybackslash}X}}
\toprule
 & Negative & Neutral & Positive \\
\midrule
Negative & 35.1 & 63.3 & 1.7 \\
Neutral  & 12.2 & 80.7 & 7.1 \\
Positive & 8.1  & 69.4 & 22.5 \\
\bottomrule
\end{tabularx}

\vspace{4pt}

\textbf{Utterance-level (N=107{,}305)}\\[-1pt]
\begin{tabularx}{\columnwidth}{l *{3}{>{\centering\arraybackslash}X}}
\toprule
 & Negative & Neutral & Positive \\
\midrule
Negative & 18.1 & 80.5 & 1.3 \\
Neutral  & 6.5  & 88.3 & 5.2 \\
Positive & 4.1  & 75.5 & 20.4 \\
\bottomrule
\end{tabularx}

\caption{Row-normalized confusion matrices (rows: NLPTown, columns: CardiffNLP). Values are percentages (rounded).}
\label{tab:conf_nlptown_card_row}
\end{table}

Standard agreement statistics contextualize inter-model classification. On the full three-way label space (\textsc{Negative}/\textsc{Neutral}/\textsc{Positive}), pairwise percent agreement and Cohen's $\kappa$ are low to moderate (Table~\ref{tab:kappa_metrics}), and overall three-model agreement measured by Fleiss' $\kappa$ is low for both granularities (Table~\ref{tab:fleiss_kappa}). When Neutral labels are included, Fleiss' $\kappa$ is 0.1287 at sentence level and 0.0535 at utterance level, confirming that inter-model disagreement is dominated by boundary decisions around neutrality rather than outright polarity reversal. When Neutral cases are excluded (polarity-only subset), Fleiss' $\kappa$ rises sharply to 0.7398 for sentences and 0.7835 for utterances (Table~\ref{tab:fleiss_kappa}), indicating that the models align much more strongly once the task is reduced to Positive-vs.-Negative polarity.

To localize the disagreement mechanism identified above, we highlight the \textsc{NLPTown}--\textsc{CardiffNLP} row-normalized confusion matrix (NLPTown rows, CardiffNLP columns), because this pair directly exposes Neutral-boundary behavior between the two three-class models. When \textsc{NLPTown} predicts \textsc{Positive}, \textsc{CardiffNLP} predicts \textsc{Neutral} in 69.4\% of sentences and 75.5\% of utterances; likewise, \textsc{NLPTown} \textsc{Negative} maps to \textsc{CardiffNLP} \textsc{Neutral} in 63.3\% (sentences) and 80.5\% (utterances) (Table~\ref{tab:conf_nlptown_card_row}). Even when \textsc{NLPTown} predicts \textsc{Neutral}, \textsc{CardiffNLP} remains \textsc{Neutral} in 80.7\% of sentences and 88.3\% of utterances (Table~\ref{tab:conf_nlptown_card_row}). Conversely, \textsc{NLPTown}'s polarity predictions frequently map to \textsc{CardiffNLP}'s \textsc{Neutral} label, directly showing that disagreement is concentrated at the Neutral boundary rather than in systematic Positive/Negative reversal.

\subsection{Stratification}

Agreement patterns are further summarized using the ABC strata (Table~\ref{tab:abc_metrics}). Unanimous polarity agreement (Category~A) is more common for sentences than utterances, whereas Category~B dominates at both granularities and Category~C remains non-trivial, indicating persistent disagreement under aggregation. Full agreement (A$_{-1}$+A$_{+1}$) covers 21.5\% of sentences but only 14.6\% of utterances, while B remains the majority at both levels (Table~\ref{tab:abc_metrics}). This makes the ABC taxonomy a practical agreement-based stability stratification for sentiment outputs in Holocaust oral histories: Category~A isolates a conservative high-consensus subset suitable for polarity-stratified sampling when higher inter-model stability is desired. Accordingly, A$_{+1}$ and A$_{-1}$ can be used as conservative polarity-specific subsets for downstream analysis, while Categories~B and~C capture the dominant disagreement region. Because A is polarity-skewed (A$_{-1}$ $>$ A$_{+1}$), polarity-stratified sampling from A should preserve this split explicitly rather than treating A as a single homogeneous ``high-agreement'' set.
\begin{figure*}[!t]
  \centering
  \includegraphics[width=\textwidth]{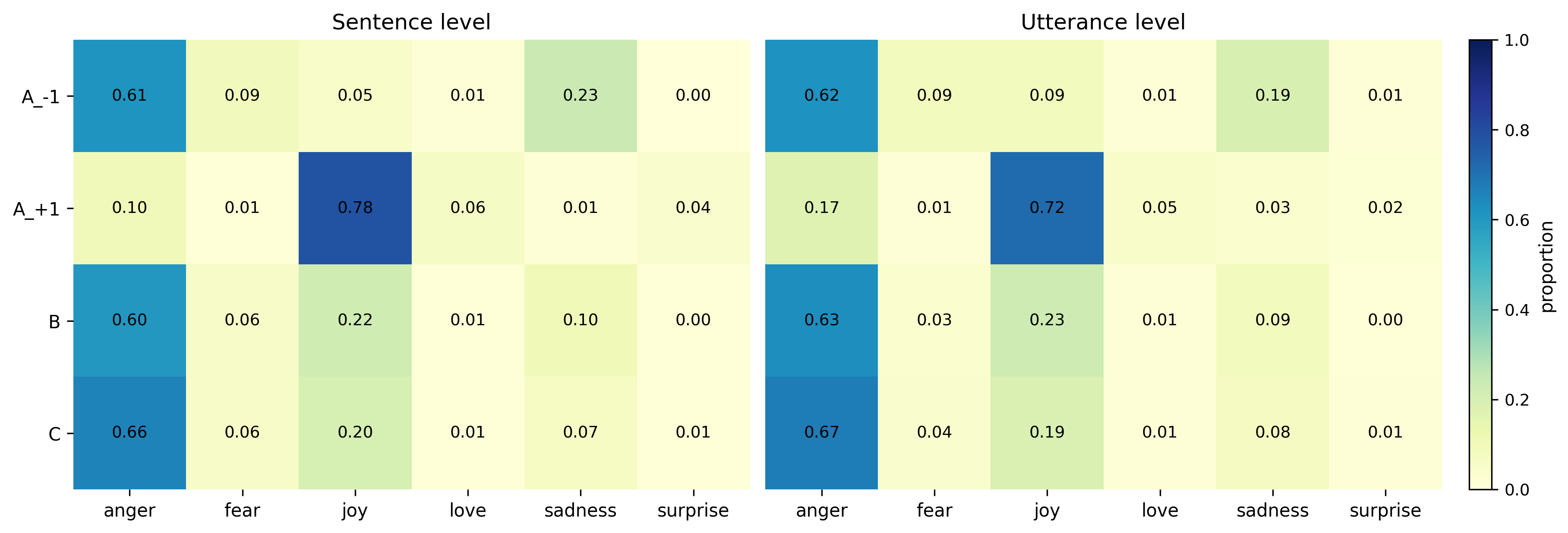}
  \caption{T5 emotion distributions (within-group percentages) across analysis groups  ($A_{-1}$, $A_{+1}$, B, C) at sentence (left) and utterance (right) levels.}
  \label{fig:emotion_dist}
\end{figure*}

\begin{figure*}[!t]
  \centering
  \includegraphics[width=\textwidth]{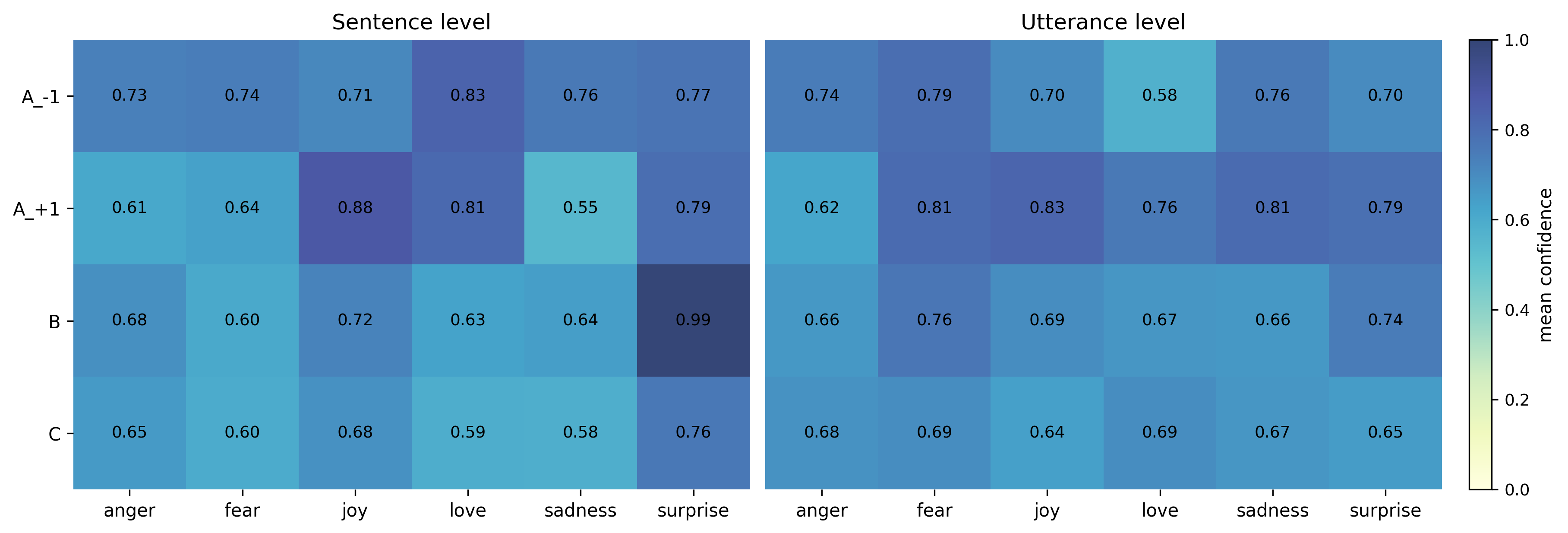}
  \caption{T5 mean confidence (uncalibrated certainty proxy) across analysis groups ($A_{-1}$, $A_{+1}$, B, C) at sentence (left) and utterance (right) levels.}
  \label{fig:emotion_conf}
\end{figure*}

\subsubsection{Descriptive Emotion Profiles}
To profile the ABC agreement strata, we apply a T5-based emotion classifier at both sentence and utterance levels. T5 is also out-of-domain in this setting; it is used only descriptively (not as validation or ground truth) to assess whether the strata exhibit coherent external affective profiles (Figures~\ref{fig:emotion_dist}--\ref{fig:emotion_conf}).

\subsubsection{Affective Distribution Patterns}
The emotion heatmaps (Figure~\ref{fig:emotion_dist}) show polarity-consistent affective profiles in the high-agreement strata.
$A_{+1}$ is dominated by joy (78\% sentence; 72\% utterance), while $A_{-1}$ is dominated by anger (61--62\%) with sadness as a substantial secondary emotion (19--23\%).
In contrast, B and C display more blended profiles (still anger-forward, with anger at 60--67\%, but with larger secondary shares of joy at 19--23\% and sadness at 7--10\%), consistent with affective heterogeneity that may contribute to cross-model polarity disagreement.

\subsubsection{T5 Confidence Proxy}
The certainty heatmaps (Figure~\ref{fig:emotion_conf}) partially mirror these patterns: $A_{+1}$ shows the highest certainty for joy (0.88 sentence; 0.83 utterance), while $A_{-1}$ shows relatively high certainty for anger (0.73 sentence; 0.74 utterance) and sadness (0.76 at both granularities).
In B and C, certainty is generally less concentrated and varies more across labels, consistent with affective ambiguity in long-form oral-history discourse, although some sparse label cells show high confidence (e.g., B--surprise at the sentence level).

\section{Conclusion}
This study shows that sentiment-classifier disagreement in Holocaust oral histories is not merely a technical nuisance but an analytically meaningful signal of domain-shift sensitivity. Rather than converging on a single sentiment profile, off-the-shelf sentiment models produce different polarity distributions at both sentence and utterance levels, with disagreement concentrated especially around the \textsc{Neutral} boundary. However, the present study is diagnostic rather than interpretive.

Triangulation and ABC provide an operational map of model behavior under domain shift: complemented by a T5-based descriptive affective probe, the framework identifies a conservative non-neutral consensus subset for downstream analysis and broader disagreement regions that can be flagged, filtered, or analyzed separately in future work.

Furthermore, these results provide a principled starting point for future work by indicating where future efforts could focus. The utterance-level aggregation rule is an operational heuristic combining within-model label frequency and confidence; alternative aggregation rules such as unweighted majority vote are left for future work. Future extensions may include domain-adaptive fine-tuning to improve polarity detection in Holocaust oral histories, introducing human-annotated evaluation subsets, and extending the ABC framework to other sensitive oral-history corpora.

\section{Ethics Statement}
This work analyzes publicly available Holocaust oral histories with respect for the survivors, their families, and the historical record. All analyses are strictly descriptive and exploratory. We do not claim that sentiment or emotion labels reflect ground-truth psychological states, nor do we present them as clinical or therapeutic interpretations. The models were used off-the-shelf (without fine-tuning on this corpus) to examine domain-shift behavior in a sensitive historical setting. Our goal is analytical: to identify where and why current NLP tools diverge on Holocaust oral histories.

\section{Acknowledgements}
I gratefully acknowledge the support of the \textit{\textbf{Deutscher Akademischer Austauschdienst (DAAD)}} through a PhD research grant (Grant No.~57645448) for my doctoral studies at \textbf{Erfurt University} (Host: \textbf{Language and Its Structure}, Prof.~Dr.~Beate Hampe). I am also grateful to Beate Hampe for reading the manuscript and providing valuable comments. I thank the anonymous reviewers for their valuable comments. Last but certainly not least, this work was carried out using \textbf{Prince}, the computational resource of the \textbf{Language and Its Structure} professorship, for which I am grateful.

\section{References}
\bibliographystyle{lrec2026-natbib}
\bibliography{lrec2026-example}
\end{document}